\documentclass{article}

\usepackage[british]{babel}
\usepackage{cite}
\usepackage{graphicx}
\usepackage{url}
\usepackage{afterpage}

\usepackage{listings}
\title{LLAMA: Leveraging Learning to Automatically Manage Algorithms}

\author{Lars Kotthoff}
\date{}

\begin{document}

\begin{titlepage}

\maketitle
\thispagestyle{empty}

\begin{center}
\includegraphics[width=.5\textwidth]{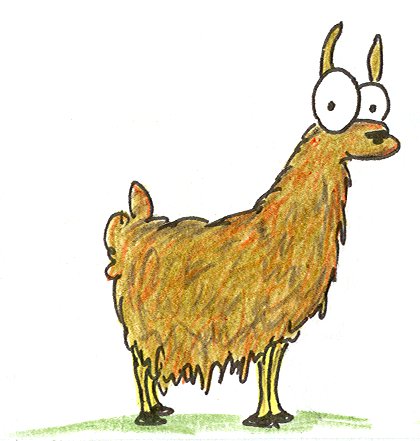}
\end{center}

\begin{abstract}
Algorithm portfolio and selection approaches have achieved remarkable
improvements over single solvers. However, the implementation of such systems is
often highly customised and specific to the problem domain. This makes it
difficult for researchers to explore different techniques for their specific
problems. We present LLAMA, a modular and extensible toolkit
implemented as an R package that facilitates the exploration of a range of
different portfolio techniques on any problem domain. It implements the
algorithm selection approaches most commonly used in the literature and
leverages the extensive library of machine learning algorithms and techniques in
R. We describe the current capabilities and limitations of the toolkit and
illustrate its usage on a set of example SAT problems.
\end{abstract}

\vspace*{\stretch{1}}

\begin{center}
This document corresponds to LLAMA version 0.6.
\end{center}

\end{titlepage}

\section*{One-page quick start}

So you know about algorithm portfolios and selection and just want to get
started. Here we go. In your R shell, type
\begin{lstlisting}
install.packages("llama")
require(llama)
\end{lstlisting}
to install and load LLAMA. We're going to assume that you have two input CSV
files for your data -- one with features and one with times. The rows designate
problem instances and the columns feature and solver names. Both files have an
``ID'' column that allows to link them. Load them into the data structure
required by LLAMA as follows.
\begin{lstlisting}
data = input(read.csv("features.csv"), read.csv("times.csv"))
\end{lstlisting}
You can also use the SAT solver data that comes with LLAMA by running
\lstinline{data(satsolvers)} followed by \lstinline{data = satsolvers}. Now
partition the entire set of instances into training and test sets for
cross-validation.
\begin{lstlisting}
folds = cvFolds(data)
\end{lstlisting}
This will give you 10 folds for cross-validation. Now we're ready to train
our first model. To do that, we'll need some machine learning algorithms --
we're going to use a random forest classifier. Load the \texttt{randomForest}
package and train a simple classification model that predicts the best
algorithm.
\begin{lstlisting}
require(randomForest)
model = classify(randomForest, folds)
\end{lstlisting}
Great! Now let's see how well this model is doing and compare its performance to
the virtual best solver (VBS) and the single best solver in terms of average
PAR10 score.
\begin{lstlisting}
mean(parscores(folds, model))
mean(parscores(data, vbs))
mean(parscores(data, singleBest))
\end{lstlisting}
You can use any other classification algorithms instead of
\lstinline{randomForest} of course. You can also train regression or cluster
models, use different train/test splits or preprocess the data by selecting the
most important features. More details in the on-line documentation, or just
continue reading for an in-depth tour of LLAMA.

\clearpage

\tableofcontents

\clearpage

\section{Background}

Throughout this document, we will assume that the reader is somewhat familiar
with algorithm portfolios, algorithm selection, and combinatorial search
problems. In this section, some of the background is explained and pointers to
additional materials given. Readers familiar with the matter may skip ahead to
the next section.

We also assume a basic familiarity with how machine learning works. Readers new
to this area can find background material in a variety of text books, e.g.\
\cite{pattern_bishop_2007,witten_data_2011,machine_lantz_2013}.

An algorithm portfolio~\cite{huberman_economics_1997,gomes_algorithm_2001} is a
collection of state of the art solvers that are all capable of solving the same
kind of problem. The rationale of using more than one algorithm or solver for a
set of problems is that no single algorithm will be the best for all of these
problems. This is known as the no free lunch theorem~\cite{wolpert_no_1997}. If
more than one solver is available, we can (at least in theory) choose the best
one for each particular problem, thus achieving superior overall performance.
The idea of algorithm portfolios was inspired by portfolios in Economics, where
a total investment is distributed over multiple securities to minimise the risk.

Many contemporary solvers for artificial intelligence problems have
complementing strengths and weaknesses. On a set of problems where one solver
exhibits bad performance, another will excel while the picture may be reversed
on a different set of problems. Algorithm portfolios exploit this by relating
the structure of the problem to solve to the performance of an individual solver
or a set of solvers.

SAT is one of the first areas of artificial intelligence that algorithm
portfolios and algorithm selection techniques have been applied to, and with
great success. The most prominent system is probably
SATzilla~\cite{xu_satzilla_2008}, which has dominated SAT solver competitions
when it was introduced. More recent systems include
ISAC~\cite{kadioglu_isac_2010}, Hydra~\cite{xu_hydra_2010} and
3S~\cite{kadioglu_algorithm_2011}.

To use an algorithm portfolio for solving problems, a selection mechanism is
required to determine the algorithm to use in the particular case. The concept
is closely related to the Algorithm Selection
Problem~\cite{rice_algorithm_1976}, which is concerned with identifying the most
suitable algorithm for solving a problem. Usually, some kind of machine learning
model is learned to relate the features of a problem instance to the performance
of an algorithm or a portfolio. Problem instance features can be anything that
describes the instance, for example structural features such as the number of
variables in a search problem or probing features such as the progress made
after running a benchmark algorithm for a short amount of time on the instance.

There are different ways in which such machine learning models can be used. In
the simplest case, a single classification model is trained to predict the best
algorithm, given the features of a problem instance. Alternatively, one
regression model per algorithm can be trained to predict its performance. The
performance predictions can then be used to choose the best algorithm. Another
approach is to cluster the problem instances in the training set, determine the
best algorithm for each cluster and assign new instances to the closest cluster.
These and many more approaches have been used in the literature. LLAMA supports
four fundamentally different approaches and a large number of variations of
these involving meta-learning techniques.

A lot more background information can be found
in~\cite{kotthoff_algorithm_2014} (even more in the extended
version~\cite{kotthoff_algorithm_2012}) and the overview table of the relevant
literature at \url{http://4c.ucc.ie/~larsko/assurvey/}.

\medskip

The main drawback of the systems described in the literature is that they are
highly tailored and customised for the particular problem domain or even set of
problems. On top of that, the implementation may not be available, or may
require an obsolete version of Matlab and the respective author's special
environment that makes it work. Even though the high-level approach can usually
be applied to other problems, in practice this is almost always very difficult
or even impossible. This makes it very difficult to compare different approaches
and prototype new ideas especially for researchers who are not algorithm
portfolio experts.

This is exactly what LLAMA addresses. Instead of providing a highly-specialised
approach that has been tuned and customised to yield high performance on a
specific data set, LLAMA is a framework that provides the building blocks for
automatic portfolio selectors. It supports the most common approaches to
portfolio selection and offers the possibility to combine them into more
sophisticated approaches. It furthermore provides an implementation of the
infrastructure that is required to build, evaluate and apply algorithm
portfolios in practice.

LLAMA is intended to be used by researchers working in the areas of algorithm
portfolios, algorithm selection, and algorithm configuration and tuning. It is
not particularly user-friendly or easy to use. It does not offer an
industrial-strength C++ implementation that you can use in a high-performance
portfolio solver. It can be used as a tool for designing such systems, but it
will not do all the required work for you.

\section{Anatomy of LLAMA}

The main focus of LLAMA is to provide the user with a framework for the
implementation and evaluation of different algorithm selection approaches. It is
\emph{not} meant to provide turn-key algorithm portfolio systems that can be
used in competitions or similar settings. While the functionality it provides
can certainly be used to facilitate the creation of such systems, a lot of the
technical details for practical algorithm selection systems are highly
domain-specific. The main audience LLAMA targets are researchers that wish
investigate and explore the performance characteristics of algorithm selection
systems in general.

\begin{figure}[htb]
\includegraphics[height=\textwidth,angle=-90]{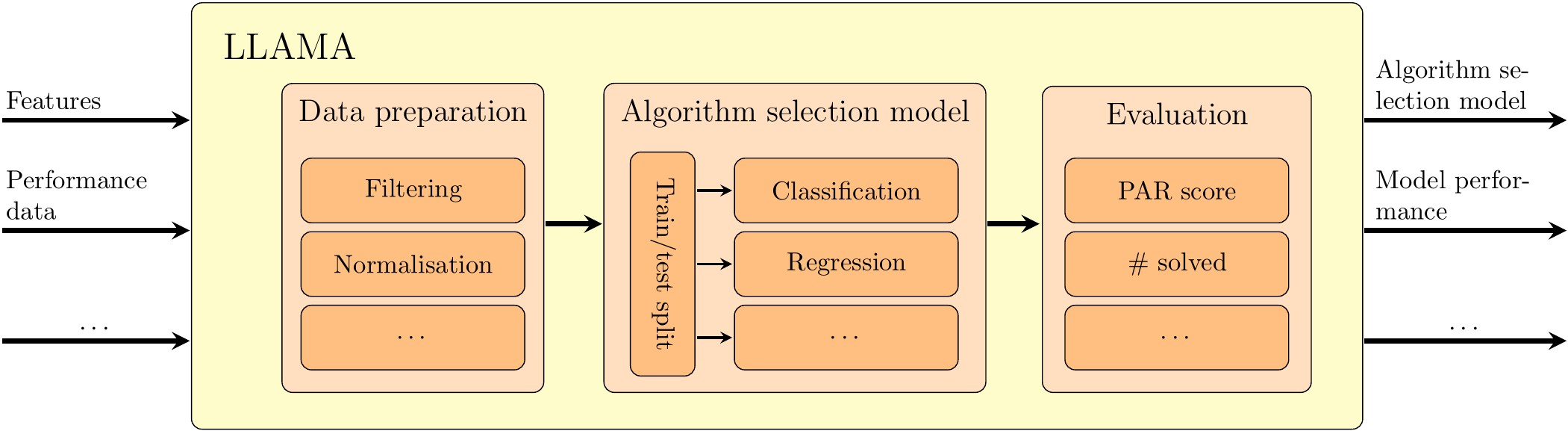}
\caption{Overview of the architecture of LLAMA.}
\label{fig:arch}
\end{figure}

The overall architecture of LLAMA is illustrated in Figure~\ref{fig:arch}. At a
high level, LLAMA takes problem feature and solver performance data as input,
processes it, and produces the algorithm selection model and a characterisation
of its performance as output. There is no explicit support for computing
features, as there are already many domain-specific systems that do this, e.g.\
SATzilla. The provided functionality falls into three main categories.

First, functions for data preparation are provided, such as filtering and
normalising the feature data, partitioning the data set for evaluation, and
analysing the contributions of the solvers in the portfolio to its overall
performance. Input data can be read from a variety of sources, e.g.\ CSV files.

The second category comprises the model-building functionality. This includes
functionality required to facilitate a clean evaluation of the learned models,
i.e.\ functions to partition into training and testing sets. All the main
approaches used in the literature are represented -- one can build models
treating algorithm selection as label classification, regression models that
predict the performance of each solver in the portfolio, clustering models that
assign the best solver to each cluster, and models that predict which solver is
faster for each pair of solvers. All this functionality is available through a
unified interface -- changing the type of algorithm selection model requires
only a different function call, changing the type of machine learning used to
induce the model requires only a change of parameter to the model-building
function call. Similarly, the output produced by these functions implements a
common interface.

The third category of functionality contains the functions used to evaluate the
learned models. Again, all commonly used evaluation measures, such as number of
instances solved and PAR10 score, are supported. These measures can be reported
for each individual problem instance or as averages.

Of the data preparation functions, any number can be used on a given data set.
One could for example read the data, normalise the feature values, and filter
the irrelevant features. In other cases, just reading the data may be
sufficient. The processed data can then be used to build one or more algorithm
selection models, depending on the requirements of the user. For a given
application, only a single model may be required, while for a performance
comparison several models would be needed. The learned models are then passed to
the evaluation functions. Again it will depend on the application whether
computing just one or several evaluation criteria makes sense.

All functions communicate through a set of common interfaces, which make it easy
to extend the functionality. To implement a new model-building approach for
example, the code to process the input and produce the output can be reused, and
the researcher is free to focus on the actual algorithm, on which no
restrictions are imposed.

\subsection{Implementation}

LLAMA is implemented as an R package. There are many advantages to this
approach; one of the main ones is that all the functionality available in R can
be used to build algorithm selection models. This is not limited to the
functionality that is implemented in R itself -- there are interfaces to many
other packages, such as the well-known Weka machine learning
toolkit~\cite{hall_weka_2009}.

The large number of machine learning approaches and algorithms available in R
makes it possible to use LLAMA to quickly evaluate a range of different
techniques for algorithm selection on given data, such as presented
in~\cite{kotthoff_evaluation_2012}. Being able to do so is crucial for achieving
good performance in practice. LLAMA has minimal requirements for machine
learning algorithm implementations it can use, allowing the user to take full
advantage of the functionality R and third-party packages provide.

\section{LLAMA for domestic use}

LLAMA is implemented as an R package and can be found at
\url{http://cran.r-project.org/web/packages/llama/}, the development repository
is at \url{https://bitbucket.org/lkotthoff/llama}. One of the main advantages of
the R package implementation is that all the functionality available in R
can be used to build performance model. This is not limited to the functionality
that is implemented in R itself -- there are interfaces to many other packages,
such as the well-known Weka machine learning toolkit~\cite{hall_weka_2009}.

The large number of machine learning approaches and algorithms available in R
makes it possible to use LLAMA to quickly evaluate a range of different
techniques for algorithm selection on given data, such as presented
in~\cite{kotthoff_evaluation_2012}. Being able to do so is crucial for achieving
good performance in practice.

LLAMA provides a number of high-level functions that can be used to create
and evaluate algorithm selection models with just a few lines of code. It is
helpful to be familiar R and its language, although this document does not
assume that you are. You will however need to be somewhat familiar with R to use
the more sophisticated functionalities of LLAMA. There are many books on R,
e.g.~\cite{rbook}.

All of the functions LLAMA provides are documented in R's online help system,
usually with examples of how to use it. To access a help page, simply type
\lstinline{?<name of function>}.

\subsection{Installing LLAMA}

LLAMA is available on CRAN. On a computer connected to the internet, all you
have to do is open an R terminal and type
\begin{lstlisting}
install.packages("llama")
\end{lstlisting}
Alternatively, you can use the graphical package manager your R distribution
provides, or download the package file yourself and install it manually.

Once the package is installed, you can load it by typing
\lstinline{require(llama)}.

\subsection{Reading data}

Let's start at the beginning -- getting your data into LLAMA. It uses a special
data structure that contains, besides the actual performance and feature data,
meta data about which algorithm was the best in which case, how to extract
feature and performance values, and other information that is required by the
various functions that operate on it. Throughout this document, we will talk
about the ``performance'' of an algorithm -- this will usually be its runtime,
but can be other things such as the quality of an obtained solution. LLAMA
places no restrictions on what ``performance'' means.

LLAMA's \lstinline{input()} function requires a particular data format, but
places no restrictions on where the data comes from. Its first argument is a
data frame that contains the features for each problem instance. Each row in the
data frame designates a different problem instance, each column holds the values
for a different feature. The second argument to \lstinline{input()} is a similar
data frame that contains performance information for the algorithms in the
portfolio. Both data frames should have a column that holds the ID of the
problem instance such that the two data frames can be merged. In fact, LLAMA
assumes that any columns that are present in both data frames can be used to
merge them.

The third (and optional) argument is a data frame that tells LLAMA whether the
run of a particular algorithm on a particular instance was successful or not.
The column names should be the same as for the data frame that holds the
algorithm performance values, and there should also be an ID column. Each cell
holds a Boolean value designating whether the run was a success or not. The
definition of ``success'' depends on the context; it can for example determine
whether an algorithm returned a solution within a certain runtime limit. If this
argument is specified, an additional way of evaluating the performance of an
algorithm selection model is available. There are no other differences; most of
the functionality of LLAMA does not require success values.

Another optional argument can be given to specify the cost of computing the
feature values. This overhead incurred by the algorithm selection system needs
to be taken into account to provide a realistic performance evaluation of the
learned models. There are three different ways of specifying feature costs. A
single number is assumed to be the cost for each instance. Alternatively, a data
frame with an ID column and a column for each feature can be given. The entries
in the rows denote the cost of computing the respective feature value for the
respective instance. The third way of specifying feature computation costs is
through a list that specifies feature groups and their costs. More details on
how to specify feature costs along with examples can be found in LLAMA's on-line
help.

If feature costs are specified, they are automatically taken into account during
the evaluation of the learned models. Cost and performance are assumed to be
additive -- that is, the cost can be added to the performance value. This covers
the most common case for algorithm selection where the performance is the
runtime. In addition to adding the cost, LLAMA also checks whether, with the
overhead included, the system would incur a timeout and takes appropriate action
if this is the case.

The final (and again optional) argument tells LLAMA whether low performance
values are good or bad. It specifies how LLAMA determines the best algorithm,
given the performances of the algorithms on an instance. The default behaviour
is to assume that smaller values are better (the values give e.g.\ runtimes).
For the opposite behaviour (e.g.\ quality of solution), specify
\lstinline{minimize=F}.

\medskip

Assume that your data is in a set of CSV files with the following format.
\begin{verbatim}
features.csv:
  ID,width,height
  0,1.2,3
  ...more instances...

performance.csv:
  ID,alg1,alg2
  0,2,5
  ...more instances...

success.csv:
  ID,alg1,alg2
  0,T,F
  ...more instances...
\end{verbatim}
You can load this data into LLAMA as follows.
\begin{lstlisting}
data = input(read.csv("features.csv"), read.csv("times.csv"), read.csv("success.csv"))
\end{lstlisting}

The \lstinline{input()} function automatically computes all the meta data
required by LLAMA -- the return value can be used right away. Full details on
the returned structure can be found in the on-line documentation.

\subsubsection{Example data}

LLAMA comes with some example data that you can play around with as a start. The
data is runtime data for 19 SAT solvers on 2433 SAT
instances~\cite{hurley_proteus_2014}. For each instance, 36 features were
measured. Success data (i.e.\ whether an algorithm timed out or not) is also
available, but feature computation costs are not.

To use this data, type
\begin{lstlisting}
data(satsolvers)
satsolvers
\end{lstlisting}
If you want to run the following commands with this data, run
\lstinline{data = satsolvers}.

\subsection{Slicing and dicing the data}

Machine learning models are usually trained and tested on separate data. This is
to avoid so-called overfitting, where the model learned is so specific to the
data it was trained on that the predictions on anything else are very
inaccurate. LLAMA provides functions to split a data set into training and test
sets. This is one of the tedious and error-prone steps that researchers have to
deal with in practice and that LLAMA aims to make less painful. To split the
data into 60\% training and 40\% test sets, we can run the following command,
assuming that your data is available in the \lstinline{data} variable.
\begin{lstlisting}
split = trainTest(data)
\end{lstlisting}
The second (optional) argument of the function specifies what fraction of the
total data should be used for training. If, instead of a 60-40 split, we want a
70-30 split, all we need to do is run \lstinline{split = trainTest(data, 0.7)}.

By default, the training and test partitions are \emph{stratified}. That is, the
distribution of best-algorithm labels in both partitions is approximately equal.
If, for example, solver A is the best on 90\% of the instances and solver B on
the remaining 10\% in the training set, the same will be the case in the test
set. To turn off this behaviour, give the additional argument
\lstinline{stratify = F}.

In addition to a simple train-test split, LLAMA also provides a function to
create data folds for cross-validation~\cite{kohavi_study_1995}, which is in
general seen as a more reliable way of evaluating the performance of a learning
algorithm. To partition the entire data into 10 stratified folds, run
\begin{lstlisting}
folds = cvFolds(satsolvers)
\end{lstlisting}
The optional argument \lstinline{nfolds} allows to specify the number of folds.
The default behaviour is again to stratify the folds; this can be turned of in
the same way as for \lstinline{trainTest()}.

\subsection{Training and evaluating models}

Now that we have both training and test data, we can train an algorithm
selection model. To start with, we will train a simple classification model
that, given the features of an instance, predicts the algorithm to use. This
approach is used for example in~\cite{gent_learning_2010}. We train a model
using the C4.5 decision tree learner~\cite{quinlan_c4.5_1993}, \lstinline{J48}
in Weka. For this, we load the \lstinline{RWeka} package and call
\lstinline{classify()} with the name of the machine learning algorithm and the
data folds created above.
\begin{lstlisting}
require(RWeka)
model = classify(J48, folds)
\end{lstlisting}

The call to \lstinline{classify} trains and tests models on each
cross-validation fold. That is, for $n$ folds it trains $n$ models using $n-1$
partitions for training and the remaining partition for testing. The predictions
on these testing partitions are returned along with a prediction function that
uses a model that was trained on the \emph{entire} data set. While the
cross-validation predictions allow to assess the expected performance of the
model, the returned prediction function can be used as a building block for a
portfolio system to obtain predictions on new data. For full details on the
returned data structure, see the on-line help.

LLAMA provides several functions to evaluate the performance of an approach
based on the predictions made. A common performance measure in the SAT community
is the PAR (penalized average runtime) score. The score is equal to the time it
took the algorithm to solve the instance or, if the algorithm was unable to
solve it, a constant factor times the time-out value. Usually, PAR10 is used,
meaning that time-outs are penalized by a factor of 10. To compute the average
PAR10 score and the total number of solved instances of the approach using the
C4.5 decision tree, we can run the following code.
\begin{lstlisting}
parscore = mean(parscores(folds, model))
solved = sum(successes(folds, model))
\end{lstlisting}
On the example data, we get 5833.845 for the mean PAR10 score and 2043 solved
instances\footnote{Note that these numbers will be slightly different when you
run the same code because the data partitioning functions are stochastic. This
also applies to all other performance numbers of models reported here}. The
\lstinline{parscores} and \lstinline{successes} evaluation
functions, along with their cousin \lstinline{misclassificationPenalties}, take
the data for which the predictions were made as their first argument and the
model that contains the predictions or the function that returns predictions (in
the case of virtual best and single best algorithms) as the second. Optional
arguments can be given to specify the penalty factor or the time-out value. By
default, the performance value given for an unsuccessful run is assumed to be
the time-out.

All evaluation functions return a list of the respective values for the chosen
algorithm for each instance. That is, if there are 100 instances in the data,
\lstinline{parscores()} will return a list with 100 values. The predictions
LLAMA computes are actually not just simple labels. Each ``prediction'' is a
data frame that contains a ranked list of algorithms along with a score value.
The number of algorithms in the data frame and the meaning of the score value
depend on the model-building function that is used and are explained in the
respective on-line help pages. In this case, each data frame will contain only a
single algorithm with a score of 1, meaning that this is the algorithm that was
predicted by the one classifier. An individual prediction looks like this.
\begin{lstlisting}
model|$\$$|predictions[[1]][[1]]
  algorithm score
1     clasp     1
\end{lstlisting}

\medskip

Computing metrics like the PAR10 score or the number of successes doesn't give
us a very good idea of how good the approach actually is. In the algorithm
selection community, two common approaches to compare against are the virtual
best solver and the single best solver. The virtual best solver assumes that we
have a perfect predictor that will always choose the best algorithm for a
particular instance. The single best solver is the algorithm in the portfolio
that has overall the best performance, i.e.\ on the largest number of instances
in the data set.

LLAMA provides convenience functions that allow to compare to both virtual best
and single best solver. They are used in the same way as the predictions from a
model are.
\begin{lstlisting}
vbsparscore = mean(parscores(data, vbs))
vbssolved = sum(successes(data, vbs))

sbparscore = mean(parscores(data, singleBest))
sbsolved = sum(successes(data, singleBest))
\end{lstlisting}
Comparing those numbers to the ones from the model that we trained should give
us a better idea of its performance. Ideally, the model performance should be
better than the one of the single best solver and as close to the virtual best
as possible. On the example data, we get a mean PAR10 score of 4645.169 and 2124
solved instances for the VBS, and a mean PAR10 score of 5779.526 and 2048 solved
instances for the best single solver. Our classification model does not improve 
on the single best solver in this case.

There are several different definitions for the single best solver, depending on
the performance measure used to determine it. The \lstinline{singleBest}
function determines it as the one that has the best cumulative performance over
all problem instances in the data set. LLAMA also provides functions to
determine the single best by PAR score (\lstinline{singleBestByPar}), by number
of problem instances solved (\lstinline{singleBestBySuccesses}), and by number
of instances it delivered the overall best performance on
(\lstinline{singleBestByCount}).

\medskip

As mentioned above, the model-building function also returns a prediction
function that allows to work with new data. As an example, we will use it to
make predictions for the data that we have. Note that we're getting predictions
for the same data that we used to train the model here -- do not use these
predictions to evaluate the performance, this example is purely to illustrate
what code to run.
\begin{lstlisting}
model|$\$$|predictor(subset(data|$\$$|data, TRUE, data|$\$$|features))
\end{lstlisting}

Oh and if you want better model performance than the single best, try the
\lstinline{randomForest} classifier from the quick start.

\subsection{Other available model types}

Instead of using the \lstinline{J48} decision tree inducer, we can use any other
classification algorithm. The only change needed is to give the other machine
learning algorithm as the first argument.

Building a classifier to predict the best solver for a problem is only one of
the approaches to providing a selector for algorithm portfolios. A different
approach is used for example in older versions of
SATzilla~\cite{xu_satzilla_2008}. For each solver in the portfolio, a regression
model is induced to predict the performance of the solver on a particular
problem. Given these predictions, the solver with the best predicted performance
is chosen.

LLAMA supports this kind of performance model as well. All we have to do is call
a different function and pass in a machine learning algorithm that is able to
learn models to predict numeric quantities as an argument. We use RWeka's
\lstinline{LinearRegression} as an example.
\begin{lstlisting}
model = regression(LinearRegression, folds)
\end{lstlisting}
You will notice that running this command takes longer than for the
classification example. This is because instead of a single classification
model, we now need to train one regression model for each algorithm in the
portfolio (if you're using the example SAT data, 19 different models).

The structure returned by the call is the same as for classification and PAR10
scores and similar are calculated in the same way. We get 5612.517 as mean PAR10
score and 2058 solved instances on the example data -- an improvement on our
earlier classification model and better than the single best algorithm. LLAMA
offers a unified interface for all its model-building functions that makes it
easy to quickly try different approaches. The difference is that the prediction
data frames now contain a row for each algorithm in the portfolio and the score
denotes the predicted performance value.
\begin{lstlisting}
model|$\$$|predictions[[1]][[1]]
        algorithm      score
1            rsat -1734.6641
2         glucose -1327.6773
3           sat4j -1262.2917
4     glueminisat -1136.2431
5       lingeling -1127.2537
6      cirminisat -1118.5151
7             mxc -1117.9649
8            riss -1110.4187
9           clasp -1109.5472
10  cryptominisat -1099.9322
11        picosat -1098.3433
12       qutersat -1084.3868
13 minisat_noelim -1062.0363
14       precosat -1054.9554
15        minisat -1020.3952
16    MPhaseSAT64 -1019.1631
17       march_rw  -968.3648
18          kcnfs  -517.6598
19      contrasat  -199.6371
\end{lstlisting}
Apart from the fact that the regression model is predicting negative runtimes,
it appears to work quite well. Negative runtimes do not matter here, as we only
use them to rank the algorithms.

The \lstinline{regression} function determines whether the lowest
performance value denotes the best algorithm by what has been specified when
running \lstinline{input}.

\medskip

The approach used in the most recent version of SATzilla is to train classifiers
that predict the better algorithm for each pair of
algorithms~\cite{xu_hydra-mip_2011}. This approach is a hybrid between the
single classification model and the regression approach. Its strength comes from
the fact that it explicitly considers the relation between two algorithms. It is
usually easier to predict which of a pair of algorithms will be better rather
than choosing the best from a large set or predicting the performance for each.

The predictions of the individual classifiers are aggregated as votes and the
algorithm that has most votes wins. The number of votes for each algorithm can
be used to rank all of the portfolio algorithms. This approach is also
implemented in LLAMA. The function is called \lstinline{classifyPairs} and
conforms to the usual interface.
\begin{lstlisting}
model = classifyPairs(J48, folds)
\end{lstlisting}
Running this command on the example SAT data will take quite a long time, as a
model for each pair of algorithms needs to be trained. This approach offers
great potential for parallelisation though; for more details, see
Section~\ref{sec:parallel}.

On the example data, we get a mean PAR10 score of 5841.625 and 2041 solved
instances. The data frame for an individual prediction now looks as follows; the
score corresponds to the number of votes.
\begin{lstlisting}
model|$\$$|predictions[[1]][[1]]
        algorithm score
1           clasp    18
2        march_rw    17
3             mxc    16
4       lingeling    15
5      cirminisat    14
6         picosat    13
7        precosat    12
8     glueminisat    11
9            rsat    10
10          kcnfs     7
11       qutersat     7
12  cryptominisat     6
13        minisat     6
14    MPhaseSAT64     6
15           riss     6
16        glucose     3
17 minisat_noelim     3
18          sat4j     1
\end{lstlisting}

\medskip

A different approach to algorithm selection that is used for example in
ISAC~\cite{kadioglu_isac_2010} is to cluster the training problems and assign
the best algorithm to each cluster based on the algorithm performances on the
instances in the cluster. Again the only change is to call a different function,
this time using RWeka's \lstinline{XMeans} clustering algorithm.
\begin{lstlisting}
model = cluster(XMeans, folds)
\end{lstlisting}
The return value corresponds to the usual format. We get a mean PAR10 score of
5736.518 and 2050 solved instances. The prediction data frames contain all
portfolio algorithms ranked by performance. The score corresponds to the sum of
the performances over all training instances in the respective cluster.
\begin{lstlisting}
model|$\$$|predictions[[1]][[1]]
        algorithm score
1         glucose  35080.60
2           sat4j  44630.40
3           clasp  71075.43
4   cryptominisat  82763.40
5         picosat  91257.26
6        qutersat  91502.81
7        precosat  92650.88
8             mxc  94293.15
9     MPhaseSAT64  94380.41
10      lingeling  94471.09
11    glueminisat  95136.49
12 minisat_noelim  97807.69
13     cirminisat  97813.11
14        minisat  97893.97
15           riss  98565.03
16      contrasat 115014.06
17       march_rw 175335.97
18           rsat 277272.10
19          kcnfs 336441.61
\end{lstlisting}

The \lstinline{cluster} model-builder provides different ways of determining the
best algorithm for a cluster that correspond to the different ways of
determining the single best algorithm. The method to use is determined by the
\lstinline{bestBy} argument, which defaults to ``performance''.

\section{Advanced functionality}

The previous section gave a glimpse of the core functionality of LLAMA. There is
much more functionality beyond that though. All the functions we have used
previously take additional arguments that allow them to be customised. The model
building functions can work with several machine learning algorithms instead of
just one. There are more functions that do exciting things\footnote{For suitable
definitions of ``exciting''.}.

\subsection{Processing the input data}

The feature and performance data is often messy -- there are missing values,
the values of some of the features are the same on all instances, or there is no
correlation between feature values and performance. All of this can impact the
performance of machine learning models. LLAMA provides functionality to mitigate
this.

\subsubsection{Selecting the most important features}

Feature selection is a process by which the features that are relevant for
making a particular prediction are identified. If, for example, the values of
one feature are the same on all problem instances, the feature does not
contribute anything and can be omitted. If on the other hand the values of a
feature change as the prediction to be made changes, we definitely want to
include this feature.

LLAMA does not offer any actual feature filtering algorithms, just like it does
not offer any machine learning algorithms. Rather, it provides the
infrastructure to use existing algorithms, which are plentiful in R. As an
example, we will use the FSelector package, which provides a number of such
functions.

Feature filtering in LLAMA is done through the \lstinline{featureFilter}
function. It takes a filtering algorithm and the data frame to use, which has to
be in the usual LLAMA format, i.e.\ what is returned by the \lstinline{input}
function.
\begin{lstlisting}
require(FSelector)
newdata = featureFilter(cfs, data)
\end{lstlisting}
The function takes the entire data and runs the feature filtering algorithm to
determine the effectiveness of the features with respect to predicting the best
algorithm. The features which are retained after filtering will be used in the
returned data frame.

This new data can now be used to split into training and test sets and build new
models.
\begin{lstlisting}
newfolds = cvFolds(newdata)

model = classify(J48, folds)
newmodel = classify(J48, newfolds)
\end{lstlisting}
The model without feature filtering achieves a mean PAR10 score of 5833.845 with
2043 instances solved, while the model with feature filtering manages a slight
improvement to a mean PAR10 score of 5645.93 and 2055 solved instances.

\subsubsection{Normalising feature values}

For some types of models, it may be desirable to normalise the feature values
such that they cover the same range over all features. If, for example, we
compute the distance between two instances based on the feature values in
Euclidean space and the values for a particular feature happen to be 1000 times
larger than the other ones, this feature will have the highest impact on the
result, even though it may not be important.

LLAMA provides a function that allows to normalise the values of features before
they are passed to the model learner. This functionality is different from the
feature filtering, which is independent from any further operations. To
normalise the feature values, scaling factors need to be computed. These same
scaling factors need to be applied when working with new data, i.e.\ when using
the predictor function returned by the model builders.

This is why normalisation is implemented as an optional argument to the
model-building functions instead of a standalone functions. The scaling factors
are computed for the training data and saved in the environment such that they
can be applied to new data later.

LLAMA currently provides only a single function, \lstinline{normalize}, for
feature value normalisation. This function scales the feature values such that
the range for all features is -1 to 1. It is specified through the
\lstinline{pre} argument.
\begin{lstlisting}
model = cluster(XMeans, folds, pre=normalize)
\end{lstlisting}
We are using the cluster model builder here, as clustering is an application
that intrinsically relies on distance measures and is likely to be most affected
by large differences in the range of feature values. The feature values can be
normalized for all of the other model-building functions in the same way.

On the example data, the mean PAR10 score is 5736.518 and the number of solved
instances 2050 -- the same as the model without normalisation. In this
particular case, normalisation does not improve the performance.

\subsubsection{Imputing censored runtimes}

Working with empirical performance data is often difficult. If the problem
instances are challenging, some of the algorithms may take a very long time to
solve them -- longer than one is prepared to wait. Usually, algorithms are run
with a time-out. That is, if the algorithm did not find a solution after a
certain amount of time, it is terminated -- its runtime is \emph{censored}.
While allows to gather data in more reasonable amounts of time, the result makes
machine learning more difficult -- if an algorithm timed out, the recorded
runtime is not actually the value we want to predict.

One way of addressing this issue is to impute the censored runtimes by learning
a machine learning model to predict the runtime on the instances that did not
time out and then apply it to the instances that timed out. This process can be
repeated to get better models and estimates~\cite{schmee_simple_1979}.

This method is implemented in LLAMA in the \lstinline{imputeCensored} function.
Its arguments are a LLAMA data frame, the regression algorithm to model the
runtime, and termination conditions. Similar to \lstinline{featureFilter}, it
returns a new data frame with the imputed censored runtimes. Note that, similar
to the regression model learner, the imputation function does not check the
plausibility of the results -- it is possible that the predicted runtimes are
less than the time-out!
\begin{lstlisting}
imputed = imputeCensored(data, LinearRegression)
imputedFolds = cvFolds(imputed)
model = regression(LinearRegression, imputedFolds)
\end{lstlisting}
The performance of the new model with imputed performance values is much better
in terms of average PAR10 and number of solved instances, but this is to be
expected -- all instances in the data are ``solvable'' after imputation, so no
penalties will be imposed. However, we can compare the mean misclassification
penalty for the old and new models. Without performance imputation, it is
88.47226 and with 45.45123 -- a clear improvement.

\subsection{Meta-learning techniques -- ensembles and stacking}

One of the main strengths of LLAMA is that meta-learning techniques can be
applied to all of the model building functions easily. The idea behind
meta-learning is similar to that of algorithm portfolios -- instead of relying
on a single machine learning algorithm and the model it learns to deliver good
predictions, we use several that (hopefully) complement each other.

The two meta-learning concepts implemented in LLAMA are
ensembles~\cite{dietterich_ensemble_2000} and
stacking~\cite{wolpert_stacked_1992}. In ensemble learning, multiple machine
learning algorithms are run on the same data, learning multiple independent
models. The predictions of each model are then combined to determine the overall
prediction. In stacking on the other hand, several machine learning algorithms
are layered on top of each other. That is, the first layer learns a model based
on the actual data, while the second layer takes the predictions of the first
layer as input. The two approaches can be combined -- the predictions of an
ensemble may be used as the input to a second layer of machine learning that
makes the final prediction.

For the \lstinline{classify} model building functions, both ensemble learning
and stacking have been implemented. To use an ensemble, the first argument
becomes a list of classification algorithms instead of a single one. To use
stacking, that list should have a member named \lstinline{.combine}.
\begin{lstlisting}
require(e1071)
ensembleModel = classify(list(J48, svm, IBk), folds)
stackedEnsembleModel = classify(list(J48, svm, IBk, .combine=J48), folds)
\end{lstlisting}
We are using the \lstinline{svm} classifier from the e1071 package in addition
to classifiers from RWeka here. The mean PAR10 scores for the ensemble and
stacked ensemble models on the example data are 5725.875 and 6394.164,
respectively. The numbers of solved instances are 2050 and 2006. Compared to the
model based on the single \lstinline{J48} classifier, ensemble learning improves
performance a bit, but stacking does not help.

The prediction data frame for ensemble models has the number of votes for each
algorithm as score. Algorithms with no votes do not appear.
\begin{lstlisting}
ensembleModel|$\$$|predictions[[1]][[1]]
  algorithm score
1     clasp     3
\end{lstlisting}

\medskip

For regression models, stacking as described in~\cite{kotthoff_hybrid_2012} is
implemented. Instead of aggregating the predicted performance values by ranking
them directly, a classifier is learned to predict, given the performance
predictions, the best algorithm. To achieve this, a classification algorithm is
given as the \lstinline{combine} argument.
set to true.
\begin{lstlisting}
stackedModel = regression(LinearRegression, folds, combine=OneR)
\end{lstlisting}
On the example data, this stacked model achieves a mean PAR10 score of 6103.86
and 2026 solved instances. The performance is worse than for the non-stacked
version with a mean PAR10 of 5612.517 and 2058 solved instances. The reason for
this is that the non-stacked combination function, choosing the algorithm with
the lowest predicted value, is very hard to learn when considering the
performance values independently and not the relation between them.

LLAMA does provide an additional argument for \lstinline{regression} that allows
to make this task easier. The \lstinline{expand} argument allows to specify a
function that, given the performance predictions, can augment the inputs to the
classifier. In this case, we want to add the pairwise absolute differences
between the predicted performances of algorithms.
\begin{lstlisting}
stackedModel = regression(LinearRegression, folds, combine=OneR,
    expand=function(x) {
        cbind(x, combn(c(1:ncol(x)), 2, function(y) {
            abs(x[,y[1]] - x[,y[2]])
        }))
    })
\end{lstlisting}
While the function given here may appear cryptic at first, it demonstrates one o
the advantages of the implementation of LLAMA as an R package. It allows to use
arbitrary R functions to process data. The mean PAR10 score achieved by this
model is 5823.698 and the number of solved instances 2044. An improvement over
the stacked model without expansion.

For the \lstinline{classifyPairs} model builder, LLAMA supports meta-learning in
the same way as for \lstinline{regression}. Stacking can be used to provide a
classification algorithm in the \lstinline{combinator} argument that learns to
predict the best algorithm, given the votes by the classifiers for pairs of algorithms.

The \lstinline{cluster} model builder supports meta-learning techniques in the
same way as \lstinline{classify}. More details and examples can be found in the
on-line documentation for each of the model builders.

\subsection{Parallel execution}\label{sec:parallel}

Some of the models can take a very long time to train. Most of the operations
are independent though and can be parallelised easily. LLAMA uses the
\lstinline{parallelMap} construct from the parallelMap
package\footnote{\url{https://github.com/berndbischl/parallelMap}} to
parallelise execution across cross-validation folds. That is, the model for each
iteration will be trained and tested in parallel. The \lstinline{parallelMap}
construct provides transparent parallelisation that executes sequentially if no
suitable parallel backend is loaded. All the user has to do to enable parallel
execution is to load a parallel backend, for example through
\lstinline{parallelStartSocket()}. Libraries used in the spawned processes
should be specified through \lstinline{parallelLibrary}.
\begin{lstlisting}
library(parallelMap)
parallelStartSocket(2) # use 2 CPUs
parallelLibrary("llama", "RWeka")
# LLAMA code
parallelStop()
\end{lstlisting}
After running these commands, all subsequent calls to LLAMA model building
functions will be parallelised across 2 CPUs.

Note that the functions provided by RWeka rely on a Java interface that is not
thread-safe. The Weka machine learning algorithms can still be used with
parallel execution though if a backend is used that runs separate processes,
such as the sockets backend used above.

\section{Case study: SATzilla}\label{sec:case:satzilla}

Algorithm selection systems often have additional components for which there is
no explicit support in LLAMA. This is not necessarily an obstacle though, much
additional functionality can be achieved by partitioning the data appropriately
and using other functionality that R provides.

This section outlines how to implement a SATzilla-like system with the help of
LLAMA and R. A full implementation is beyond the scope of this document;
instead, this section serves as a guide to researchers wishing to implement such
measures. There are several techniques that the various versions of SATzilla use
to make algorithm portfolios more performance in practice. We will sketch the
implementation of each in turn before putting it all together. The techniques
used in this section are described in more details in the SATzilla papers
(e.g.~\cite{xu_satzilla_2008,xu_hierarchical_2007,xu_hydra-mip_2011}).

The main difference of the implementation sketched here and SATzilla is that we
treat all problem instances the same way, without computing any weights that
determine the ``importance'' of the problem instance. In practice, using weights
is almost always a good idea.

\subsection{Presolver}

There are problem instances that can be solved almost instantaneously. For
these, the overhead of computing features, running the algorithm selection
model, and selecting a solver is too high. Using a portfolio is always going to
be slower for these problems. To avoid this issue, we can run a presolver for a
short amount of time.

This step is largely independent of training algorithm selection models. We do
however need to take into account that the problems that are solved by the
presolver within its time limit do not require algorithms to be selected for
them, so we should remove them from the data before training the model. This is
done easily in R.
\begin{lstlisting}
performanceData = read.csv("performance.csv")
presolveLimit = 1
presolver = "minisat"
newPerformanceData = subset(performanceData, performanceData[presolver] > presolveLimit)
data = input(read.csv("features.csv"), newPerformanceData)
\end{lstlisting}
Assuming that our presolver is minisat with a time-out of 1 second, we can
filter the data as above before passing it to \lstinline{input}. There is no
need to filter the feature data, as LLAMA will discard any items it cannot
match. The resulting LLAMA data frame can be used as usual to train and evaluate
models.

\subsection{Prediction of satisfiability}

SAT solvers often behave differently depending on whether an instance is
satisfiable or not. In practice, it can make sense to distinguish between these
instances in algorithm portfolios. This requires a larger number of machine
learning models. First, we need to be able to predict whether a given instance
is satisfiable or not. Second, we need different algorithm selection models for
the satisfiable and unsatisfiable cases.

To achieve this, we require additional data on whether an instance is
satisfiable or not.
\begin{lstlisting}
performanceData = read.csv("performance.csv")
featureData = read.csv("features.csv")
featureNames = names(featureData)[-1]
satisfiable = read.csv("satisfiable.csv")
satisfiableModel = J48(satisfiable|$\$$|satisfiable~., data=subset(featureData, T, featureNames))
satisfiableData = input(featureData, subset(performanceData, satisfiable|$\$$|satisfiable))
unsatisfiableData = input(featureData, subset(performanceData, !satisfiable|$\$$|satisfiable))
\end{lstlisting}
For simplicity, we assume that the first column of the feature file is an ID and
that the order of the instances in the feature, performance, and satisfiable
files is the same. The \lstinline{satisfiableModel} can now be used to predict
the satisfiability of a new instance, while the LLAMA data frames
\lstinline{satisfiableData} and \lstinline{unsatisfiableData} can be used to
train and evaluate algorithm selection models for the respective parts of the
instance space.

\subsection{Prediction of feature computation time}

Computing the features of an instance is integral to algorithm selection
systems. This information can be related to the performance of the algorithms to
make predictions as to which algorithm to use in a particular case. However,
sometimes just computing the features can take longer than it would take to
solve the instance. It is clearly desirable to identify these cases before
starting to compute the features.

To achieve this, we can train yet another machine learning model that, given a
small and cheaply computable subset of the features, predicts the time required
to compute the remaining features. If the time is too high, we simply run a
backup solver on the instance. Such instances should not be used in the training
and evaluation of the algorithm selection model.
\begin{lstlisting}
featureTimes = read.csv("featureTimes.csv")
reducedFeatureData = read.csv("reducedFeatures.csv")
reducedFeatureNames = names(reducedFeatureData)[-1]
featureTimeModel = LinearRegression(featureTimes|$\$$|time~., data=subset(reducedFeatureData, T, reducedFeatureNames))
featureTimeLimit = 5
data = subset(subset(read.csv("features.csv"), featureTimes|$\$$|time < featureTimeLimit), read.csv("performance.csv"))
\end{lstlisting}
Similar assumptions as in the section above are made. The LLAMA data frame can
now again be used in the usual manner to train algorithm selection models.

\subsection{Putting it all together}

The techniques outlined in the previous section can now be put together to
implement a system similar to SATzilla.
\begin{lstlisting}
performanceData = read.csv("performance.csv")
featureData = read.csv("features.csv")
featureNames = names(featureData)[-1]
reducedFeatureData = read.csv("reducedFeatures.csv")
reducedFeatureNames = names(reducedFeatureData)[-1]
featureTimes = read.csv("featureTimes.csv")
satisfiable = read.csv("satisfiable.csv")

presolveLimit = 1
presolver = "minisat"
featureTimeLimit = 5

toSolve = (performanceData[presolver] > presolveLimit)&(featureTimes|$\$$|time < featureTimeLimit)
featureTimeModel = LinearRegression(featureTimes|$\$$|time~., data=subset(reducedFeatureData, T, reducedFeatureNames))
satisfiableModel = J48(satisfiable|$\$$|satisfiable~., data=subset(featureData, toSolve, featureNames))

satisfiableData = input(featureData, subset(performanceData, satisfiable|$\$$|satisfiable&toSolve))
unsatisfiableData = input(featureData, subset(performanceData, (!satisfiable|$\$$|satisfiable)&toSolve))

satisfiableFolds = cvFolds(satisfiableData)
unsatisfiableFolds = cvFolds(unsatisfiableData)
satisfiableModel = classifyPairs(J48, satisfiableFolds)
unsatisfiableModel = classifyPairs(J48, unsatisfiableFolds)
\end{lstlisting}
Let's walk through it step by step. First, we read all the relevant data, as in
the individual steps above. Then we set our time limits and the presolver. The
data we want to train algorithm selection models for comprises the instances
that are not solved by the presolver and for which the feature prediction time
is not too long. We save the Boolean mask that encodes these conditions in
\lstinline{toSolve} to be able to filter the data later.

After that, we train our first two models -- one to predict the feature
computation time and the other to predict whether an instance is satisfiable or
not. For the feature computation time, we need to consider all the data, whereas
for the prediction of satisfiability we only need the instances that we want
algorithm selection models for.

Then, we partition the data based on satisfiability and pass them to
\lstinline{include}, filtering the instances that we do not require algorithm
selection for. Finally, we proceed in the usual LLAMA manner by partitioning the
data into folds and learning a SATzilla 2012-style pairwise classification
model.

To make predictions on new data, we would use the two preliminary models to
determine feature computation time and satisfiability and then the
\lstinline{predictor} member of either the \lstinline{satisfiableModel} or the
\lstinline{unsatisfiableModel}.

Our SATzilla-style solver now looks something like this.
\begin{lstlisting}
function myPortfolioSolver(instance) {
    presolver = "minisat"
    presolveTimeout = 1
    backupsolver = "clasp"
    featureTimeout = 5

    result = runSolver(instance, presolver, timeout=presolveTimeout)
    if(!isSolved(result)) {
        simpleFeatures = getSimpleFeatures(instance)
        featureTime = predict(featureTimeMode, simpleFeatures)
        if((featureTime > featureTimeLimit) {
            return(runSolver(instance, backupsolver))
        } else {
            features = getFeatures(instance)
            satisfiable = predict(satisfiableModel, features)
            if(satisfiable) {
                solver = satisfiableModel|$\$$|predictor(features)
            } else {
                solver = unsatisfiableModel|$\$$|predictor(features)
            }
            return(runSolver(instance, solver))
        }
    } else {
        return(result)
    }
}
\end{lstlisting}

\section{Case study: Visualising the data}\label{sec:case:vis}

Our final case study is going to be a bit more graphic than the previous rather
dry sections -- we are going to have a look at visualising the data we have been
working with. R offers many possibilities for doing so and surveying them all
is far beyond the scope of this manual. We only give a flavour of what
visualisations can be done.

\begin{figure}[htb]
\begin{center}
\includegraphics[width=.9\textwidth]{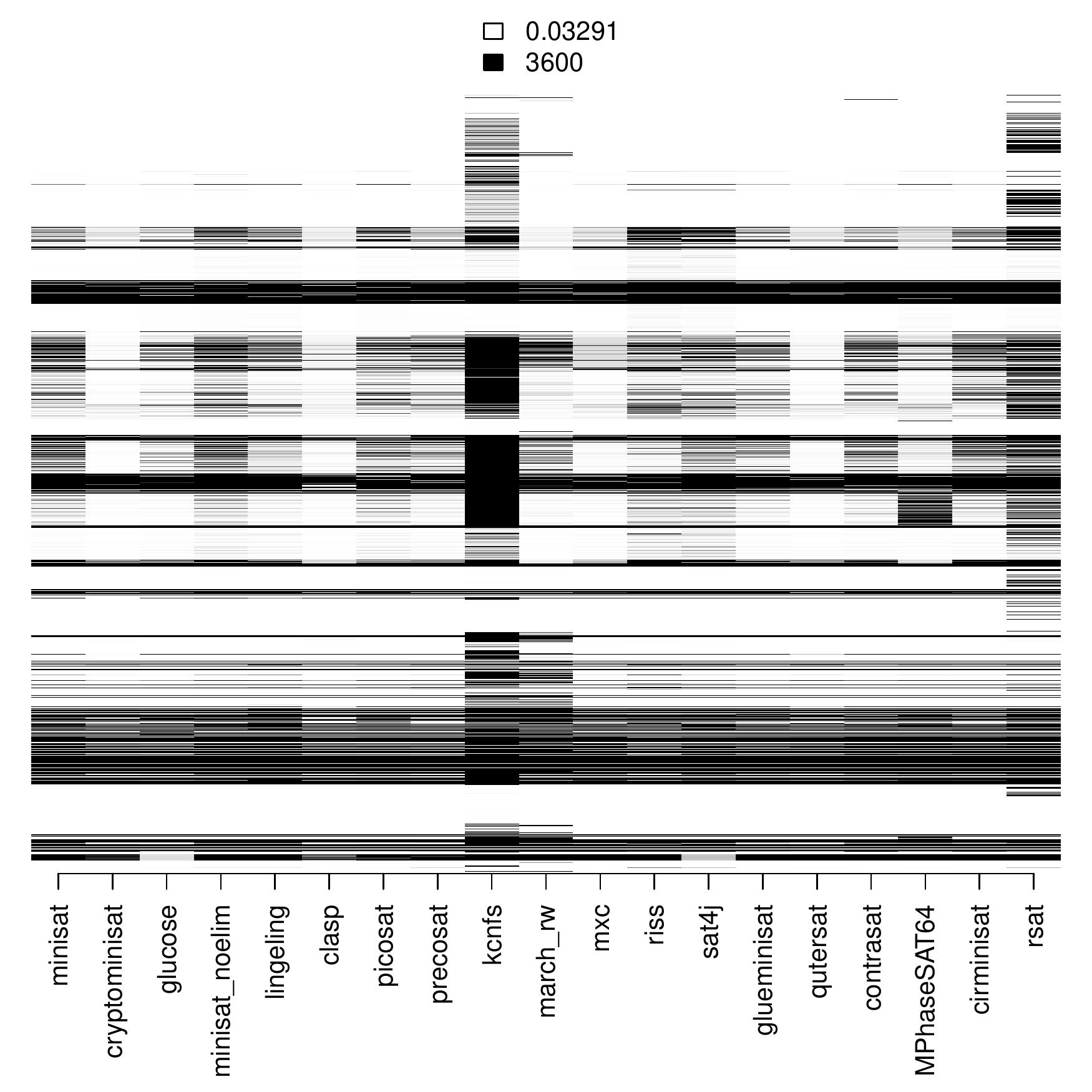}
\caption{Runtimes by solver and instance from almost instantaneous solve (white)
to timeout (black).}
\label{fig:times}
\end{center}
\end{figure}

It is always useful to have a look at the data one will be working with. In our
case, there are two main groups of data -- the performance values and the
features. Let's have a look at the performance values first. Throughout this
section, we will be using the example SAT data, but the same methodology applies
to any other data in LLAMA format. As we have a large amount of data to
visualise, we are going to use heatmaps.
\begin{lstlisting}
data(satsolvers)
times = subset(satsolvers|$\$$|data, T, satsolvers|$\$$|performance)
par(mar=c(7,1,3,1))
cols = gray(seq(1, 0, length.out=255))
image(t(as.matrix(times)), axes=F, col=cols)
axis(1, labels=satsolvers|$\$$|performance, at=seq(0, 1, 1/(length(satsolvers|$\$$|performance)-1)), las=2)
legend("top", legend=c(min(times), max(times)), fill=c("white", "black"), bty="n", inset=-0.12, xpd=NA)
\end{lstlisting}
The graph created by this code is shown in Figure~\ref{fig:times}. After
extracting the performance values (runtimes in this case) from the LLAMA data
frame, setting some plot parameters and creating a black and white colour scale,
we plot the heatmap, axis, and legend. The map shows the time each solver takes
on each instance.

\begin{figure}[htb]
\begin{center}
\includegraphics[width=.9\textwidth]{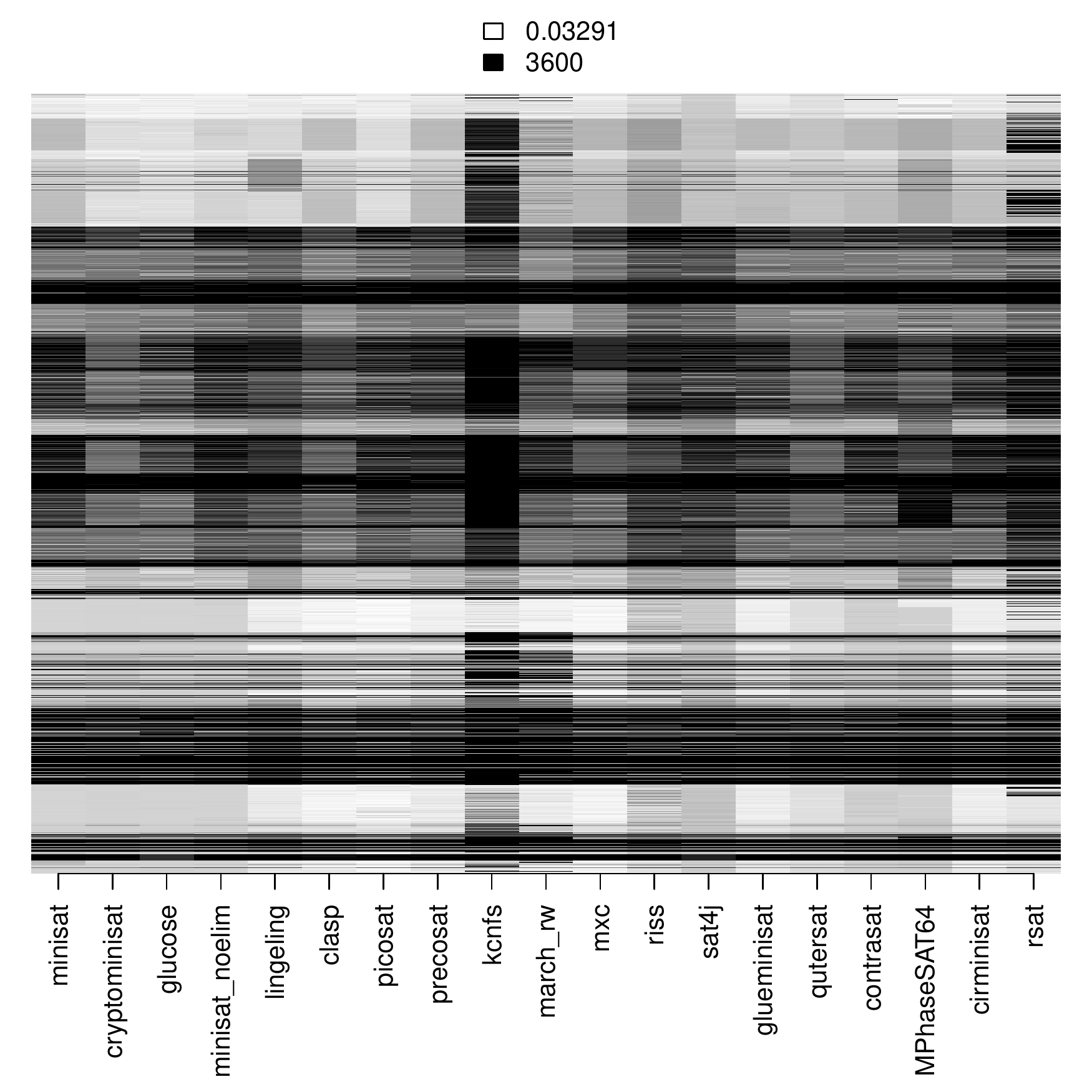}
\caption{Log runtimes by solver and instance from almost instantaneous solve
(white) to timeout (black).}
\label{fig:times-log}
\end{center}
\end{figure}

There is a large spread of runtimes in our data, with most instances being
solved either instantaneously or timing out. The figure looks accordingly, with
most colours being either white or black and very little grey in between.
Plotting the log of the runtime is more informative
(Figure~\ref{fig:times-log}).
\begin{lstlisting}
image(log10(t(as.matrix(times))), axes=F, col=cols)
axis(1, labels=satsolvers|$\$$|performance, at=seq(0, 1, 1/(length(satsolvers|$\$$|performance)-1)), las=2)
legend("top", legend=c(min(times), max(times)), fill=c("white", "black"), bty="n", inset=-0.12, xpd=NA)
\end{lstlisting}

\begin{figure}[htb]
\begin{center}
\includegraphics[width=.9\textwidth]{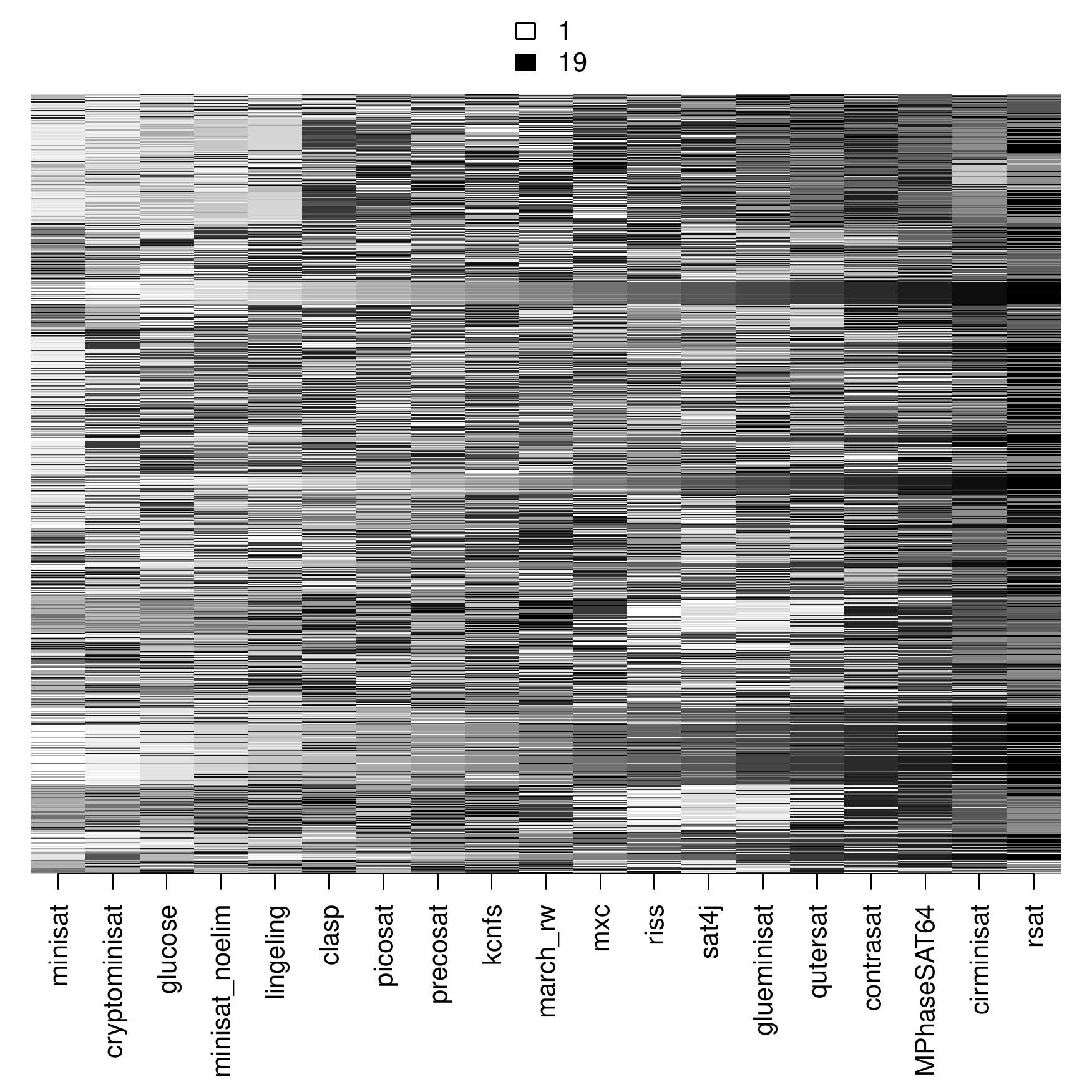}
\caption{Solver ranks on each instance from first (white) to last (black).}
\label{fig:rank}
\end{center}
\end{figure}

When doing algorithm selection, we are usually more interested in how the
performance of a solver compares to the other solvers in the portfolio rather
than the absolute value. We can plot the rank of each solver on each instance in
a similar fashion to before. The resulting plot is shown in
Figure~\ref{fig:rank}.
\begin{lstlisting}
image(apply(times, 1, order), axes=F, col=cols)
axis(1, labels=satsolvers|$\$$|performance, at=seq(0, 1, 1/(length(satsolvers|$\$$|performance)-1)), las=2)
legend("top", legend=c(1, length(satsolvers|$\$$|performance)), fill=c("white", "black"), bty="n", inset=-0.12, xpd=NA)
\end{lstlisting}

\medskip

\begin{figure}[htb]
\begin{center}
\includegraphics[width=.9\textwidth]{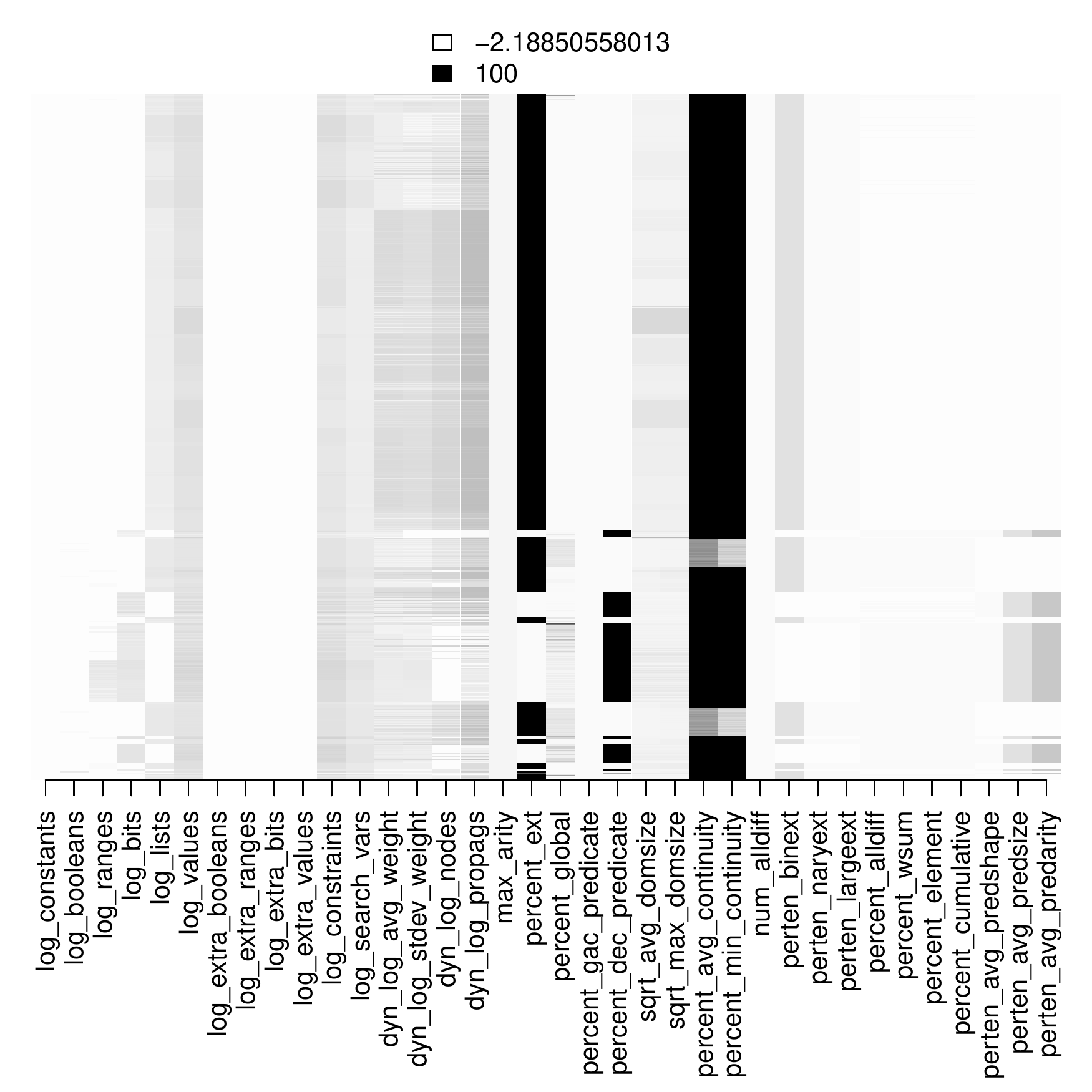}
\caption{Feature values on each instance from lowest (white) to highest
(black).}
\label{fig:features}
\end{center}
\end{figure}

The same kind of visual analysis can be applied to the features. This is
probably even more important that analysing the performance data, as the
features are used to create the models we are going to use for algorithm
selection. If the features are bad, the models will not be good either. Let's do
the same kind of plot we did for the solvers for the features.
\begin{lstlisting}
features = subset(satsolvers|$\$$|data, T, satsolvers|$\$$|features)
par(mar=c(10,1,3,1))
image(t(as.matrix(features)), axes=F, col=cols)
axis(1, labels=satsolvers|$\$$|features, at=seq(0, 1, 1/(length(satsolvers|$\$$|features)-1)), las=2)
legend("top", legend=c(min(features), max(features)), fill=c("white", "black"), bty="n", inset=-0.12, xpd=NA)
\end{lstlisting}
The resulting plot is shown in Figure~\ref{fig:features}. The values for four of
the features are much higher than the rest. It is hard to see the variance of
the feature values for a particular feature over the set of instances because of
this.

\begin{figure}[htb]
\begin{center}
\includegraphics[width=.9\textwidth]{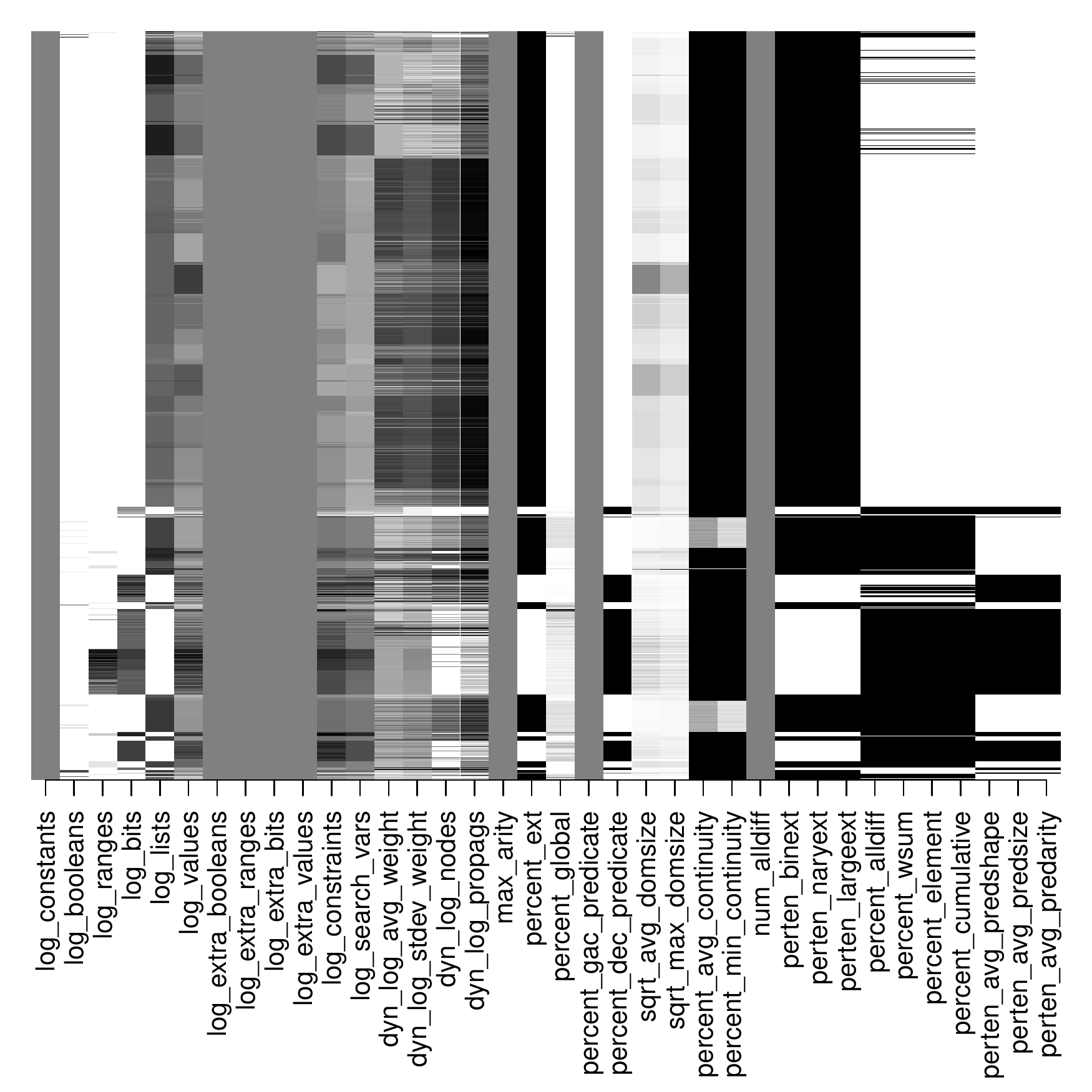}
\caption{Normalised feature values on each instance from -1 (white) to 1
(black).}
\label{fig:features-norm}
\end{center}
\end{figure}

We get a better plot from the normalised feature values
(Figure~\ref{fig:features-norm}).
\begin{lstlisting}
nFeatures = normalize(features)
par(mar=c(10,1,1,1))
image(t(as.matrix(nFeatures|$\$$|features)), axes=F, col=cols)
axis(1, labels=satsolvers|$\$$|features, at=seq(0, 1, 1/(length(satsolvers|$\$$|features)-1)), las=2)
\end{lstlisting}
The features for the example data are not very good. For quite a lot of them,
there is almost no variation and others have only two different values.

\begin{figure}[htb]
\begin{center}
\includegraphics[width=.9\textwidth]{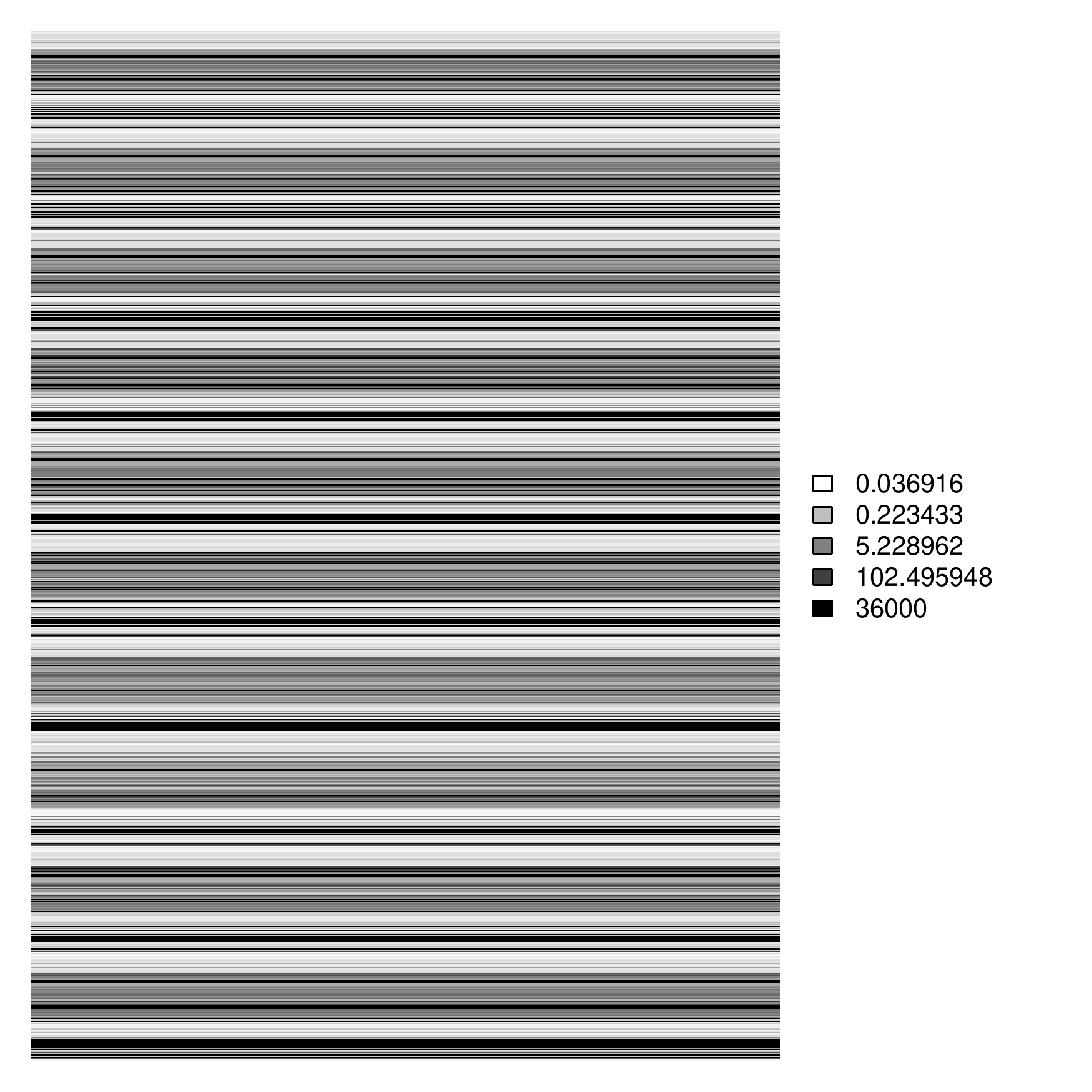}
\caption{PAR10 scores for a \texttt{J48} classification model.}
\label{fig:classification}
\end{center}
\end{figure}

We can apply the same kind of analysis to the prediction results. This allows us
to get an idea of what the machine learning algorithms are doing, where they
work well and where things could be improved. The following code creates a
heatmap of the PAR10 scores of a classification model on all instances. We are
using the log of the PAR10 scores to get more shades of grey.
\begin{lstlisting}
folds = cvFolds(satsolvers)
model = classify(J48, folds)
scores = parscores(folds, model)
par(mar=c(1,1,1,10))
image(log10(t(as.matrix(scores))), axes=F, col=cols)
legend("right", legend=quantile(scores), fill=gray(seq(1, 0, length.out=5)), bty="n", inset=-0.3, xpd=NA)
\end{lstlisting}
The resulting plot is shown in Figure~\ref{fig:classification}. It gives us some
idea of what's going on, but looks more like a bar code than anything else.

\begin{figure}[htb]
\begin{center}
\includegraphics[width=.9\textwidth]{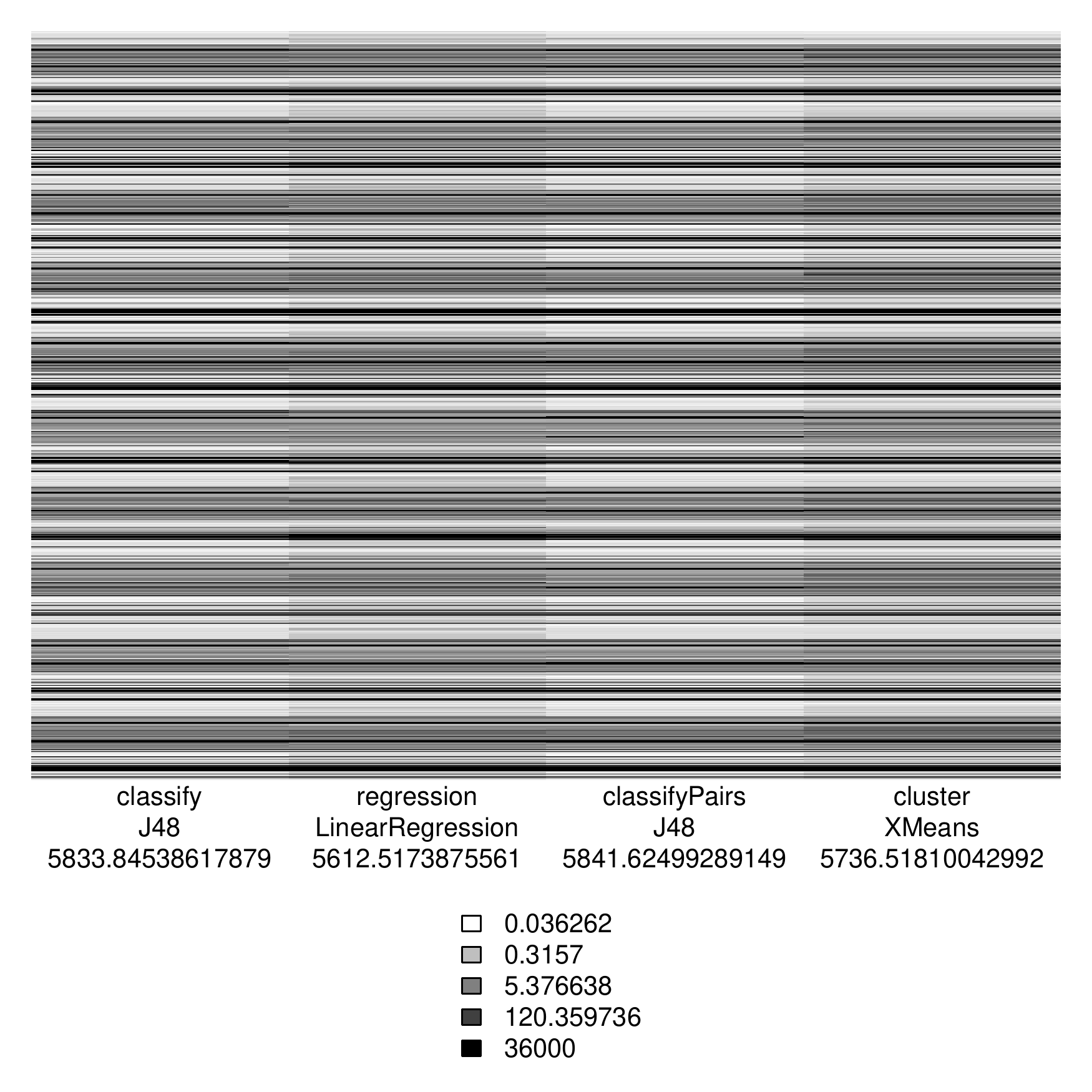}
\caption{PAR10 scores for four different models. The number in each label shows
the mean PAR10 score across all instances.}
\label{fig:comparison}
\end{center}
\end{figure}

The power of heatmaps comes from being able to compare multiple entities. A more
useful plot compares the performance of the classification model to the
performance of other models. The following code can be used to generate such a
plot (Figure~\ref{fig:comparison}).
\begin{lstlisting}
folds = cvFolds(satsolvers)
model1 = classify(J48, folds)
model2 = regression(LinearRegression, folds)
model3 = classifyPairs(J48, folds)
model4 = cluster(XMeans, folds)
scores1 = parscores(folds, model1)
scores2 = parscores(folds, model2)
scores3 = parscores(folds, model3)
scores4 = parscores(folds, model4)
par(mar=c(10,1,1,1))
names = c(paste("classify", "J48", mean(scores1), sep="\n"), paste("regression", "LinearRegression", mean(scores2), sep="\n"), paste("classifyPairs", "J48", mean(scores3), sep="\n"), paste("cluster", "XMeans", mean(scores4), sep="\n"))
scores = c(scores1, scores2, scores3, scores4)
image(log(t(matrix(scores, ncol=4))), axes=F, col=cols)
axis(1, labels=names, at=seq(0, 1, 1/(length(names)-1)), las=1, tick=F, line=1)
legend("bottom", legend=quantile(scores), fill=gray(seq(1, 0, length.out=5)), bty="n", inset=-0.4, xpd=NA)
\end{lstlisting}
This comparison is actually quite insightful. The predictions of the regression
model usually incur a higher PAR10 score (overall, the column is darker than the
others), but its mean PAR10 score is lower than for the other models. It also
shows us that for this specific scenario, there is almost no difference between
the single and pairwise classification models.

\begin{figure}[htb]
\begin{center}
\includegraphics[width=.9\textwidth]{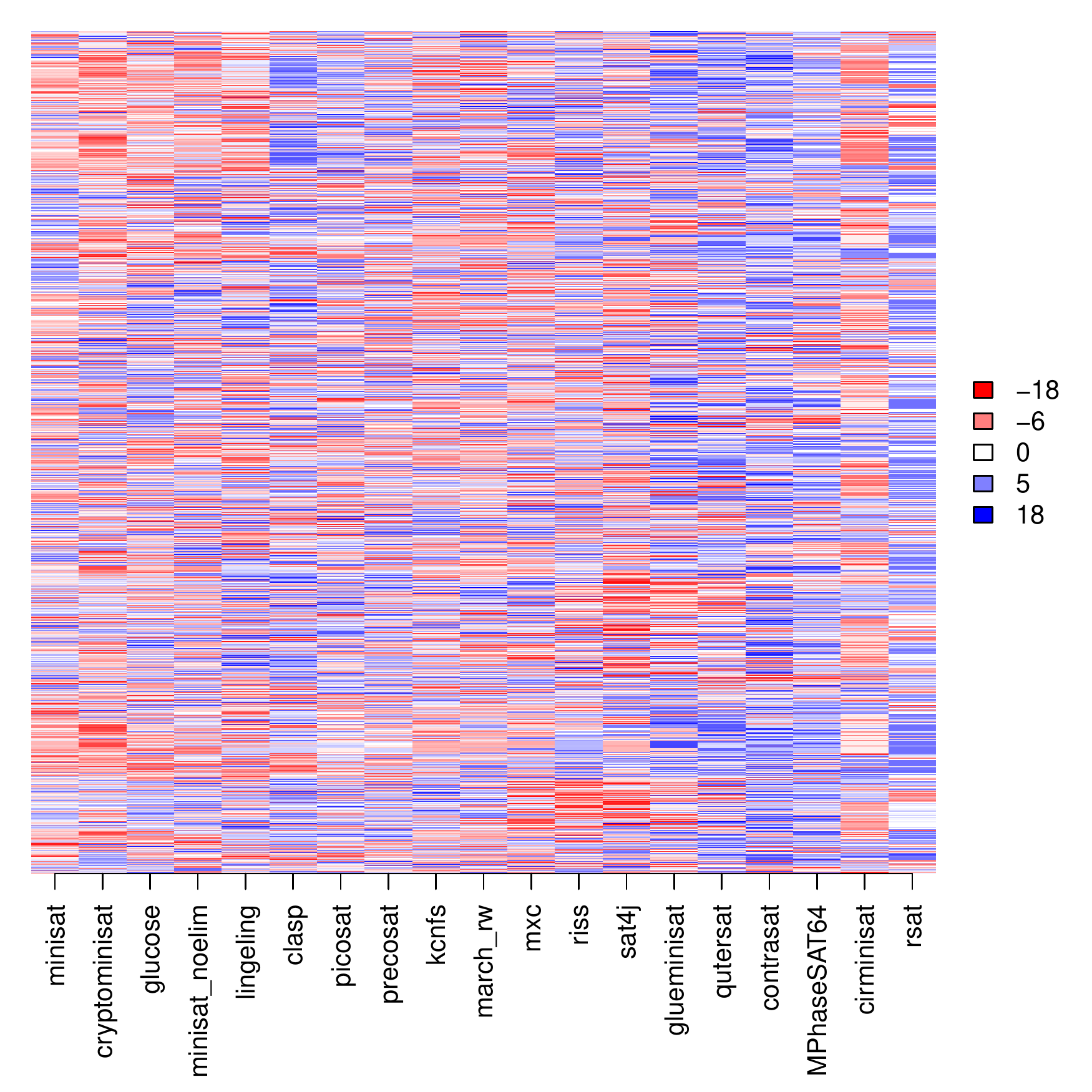}
\caption{Difference between actual and predicted rank for the regression model.
Red means that an algorithm was predicted to be worse than it actually was, blue
that it was predicted to be better. Predicted and actual rank agree for white.}
\label{fig:rank-diff}
\end{center}
\end{figure}

As a final example, we will plot the differences in predicted and actual rank
for the regression model. The heatmap resulting from the code below is shown in
Figure~\ref{fig:rank-diff}.
\begin{lstlisting}
par(mar=c(7,1,1,5))
ranks = apply(sapply(unlist(model2|$\$$|predictions, recursive=F), function(x) { x|$\$$|algorithm }), 2, function(y) { as.vector(sapply(y, function(z) { which(z == satsolvers|$\$$|performance) })) })
diffs = apply(times, 1, order)-ranks
cols = c(colorRampPalette(c("red", "white"))(128), colorRampPalette(c("white", "blue"))(128))
breaks = c(seq(min(diffs), -1/128, length.out=128), 0, seq(1/128, max(diffs), length.out=128))
image(diffs, axes=F, col=cols, breaks=breaks)
axis(1, labels=satsolvers|$\$$|performance, at=seq(0, 1, 1/(length(satsolvers|$\$$|performance)-1)), las=2)
legend("right", legend=quantile(diffs), fill=cols[quantile(1:length(cols))], bty="n", inset=-0.15, xpd=NA)
\end{lstlisting}

The plot is a bit messy, but there are some insights to be gained. Looking at
the solvers that perform well overall (e.g.\ minisat, leftmost column), we can
see that they are predominantly red, i.e.\ most of the time they are predicted
to be worse than they actually are. For the solvers that exhibit bad performance
(e.g.\ rsat, rightmost column), we observe that they are predominantly blue,
meaning that they are usually predicted to be better than they actually are.
This indicates that improvements are to be had by training better machine
learning models in this particular scenario.

\medskip

The examples above only scratch the surface of the visualisations that are
possible in R. While none of the examples shown depends on LLAMA, the data
structures it provides facilitate the creation of such plots to some extent.

\section{The further domestication of LLAMA}

The functionality currently implemented in LLAMA facilitates the exploration of
the performance of many different approaches to algorithm selection. It is our
intention for LLAMA to develop into a platform that not only facilitates the
exploration and comparison of different existing approaches, but also the rapid
prototyping of new approaches. The functions it provides take care of the
infrastructure required to train and evaluate machine learning models in a
scientifically rigorous way. Researchers wishing to improve algorithm portfolio
selection do not need to concern themselves with the infrastructure, but can
concentrate on the actual research.

LLAMA is still at an early stage of development and bugs may occur. In the end,
the responsibility of interpreting and validating the results lies with the
user. However, we have used LLAMA for a number of applications, found and fixed
a few bugs and are reasonably confident that it will useful to other people.

\subsection*{Acknowledgements}

This work is supported by EU FP7 ICT-FET grant 284715 (ICON). The drawing of a
Llama is courtesy of \url{http://www.bluebison.net/}. We wish to thank all the
people who contributed to LLAMA.

\bibliography{\jobname}

\begin{thebibliography}{10}

\bibitem{pattern_bishop_2007}
Christopher~M. Bishop.
\newblock {\em Pattern Recognition and Machine Learning}.
\newblock Springer, 2007.

\bibitem{rbook}
Michael~J. Crawley.
\newblock {\em The R Book}.
\newblock Wiley Publishing, 1st edition, 2007.

\bibitem{dietterich_ensemble_2000}
Thomas~G. Dietterich.
\newblock Ensemble methods in machine learning.
\newblock In {\em International Workshop on Multiple Classifier Systems}, pages
  1--15, 2000.

\bibitem{gent_learning_2010}
Ian~P. Gent, Christopher~A. Jefferson, Lars Kotthoff, Ian Miguel, Neil Moore,
  Peter Nightingale, and Karen~E. Petrie.
\newblock Learning when to use lazy learning in constraint solving.
\newblock In {\em 19th European Conference on Artificial Intelligence}, pages
  873--878, August 2010.

\bibitem{gomes_algorithm_2001}
Carla~P. Gomes and Bart Selman.
\newblock Algorithm portfolios.
\newblock {\em Artificial Intelligence}, 126(1-2):43--62, 2001.

\bibitem{hall_weka_2009}
Mark Hall, Eibe Frank, Geoffrey Holmes, Bernhard Pfahringer, Peter Reutemann,
  and Ian~H. Witten.
\newblock The {WEKA} data mining software: An update.
\newblock {\em {SIGKDD} Explor. Newsl.}, 11(1):10--18, November 2009.

\bibitem{huberman_economics_1997}
Bernardo~A. Huberman, Rajan~M. Lukose, and Tad Hogg.
\newblock An economics approach to hard computational problems.
\newblock {\em Science}, 275(5296):51--54, 1997.

\bibitem{hurley_proteus_2014}
Barry Hurley, Lars Kotthoff, Yuri Malitsky, and Barry {O'Sullivan}.
\newblock Proteus: A hierarchical portfolio of solvers and transformations.
\newblock In {\em {CPAIOR}}, May 2014.

\bibitem{kadioglu_algorithm_2011}
Serdar Kadioglu, Yuri Malitsky, Ashish Sabharwal, Horst Samulowitz, and Meinolf
  Sellmann.
\newblock Algorithm selection and scheduling.
\newblock In {\em 17th International Conference on Principles and Practice of
  Constraint Programming}, pages 454--469, 2011.

\bibitem{kadioglu_isac_2010}
Serdar Kadioglu, Yuri Malitsky, Meinolf Sellmann, and Kevin Tierney.
\newblock {ISAC} – instance-specific algorithm configuration.
\newblock In {\em Proceeding of the 2010 conference on {ECAI} 2010: 19th
  European Conference on Artificial Intelligence}, pages 751--756, Amsterdam,
  The Netherlands, The Netherlands, 2010. {IOS} Press.

\bibitem{kohavi_study_1995}
Ron Kohavi.
\newblock A study of cross-validation and bootstrap for accuracy estimation and
  model selection.
\newblock In {\em Proceedings of the 14th International Joint Conference on
  Artificial Intelligence}, pages 1137--1143. Morgan Kaufmann, 1995.

\bibitem{kotthoff_algorithm_2012}
Lars Kotthoff.
\newblock Algorithm selection for combinatorial search problems: A survey.
\newblock Technical Report {arXiv:1210.7959}, University College Cork, 2012.

\bibitem{kotthoff_hybrid_2012}
Lars Kotthoff.
\newblock Hybrid regression-classification models for algorithm selection.
\newblock In {\em 20th European Conference on Artificial Intelligence}, pages
  480--485, August 2012.

\bibitem{kotthoff_algorithm_2014}
Lars Kotthoff.
\newblock Algorithm selection for combinatorial search problems: A survey.
\newblock {\em {AI} Magazine}, 2014.
\newblock Forthcoming.

\bibitem{kotthoff_evaluation_2012}
Lars Kotthoff, Ian~P. Gent, and Ian Miguel.
\newblock An evaluation of machine learning in algorithm selection for search
  problems.
\newblock {\em {AI} Communications}, 25(3):257--270, 2012.

\bibitem{machine_lantz_2013}
Brett Lantz.
\newblock {\em Machine Learning with R}.
\newblock Packt, 2013.

\bibitem{quinlan_c4.5_1993}
J.~Ross Quinlan.
\newblock {\em C4.5: Programs for Machine Learning}.
\newblock Morgan Kaufmann, 1 edition, January 1993.

\bibitem{rahwan_game_2013}
Talal Rahwan and Tomasz~P. Michalak.
\newblock A game theoretic approach to measure contributions in algorithm
  portfolios.
\newblock Technical Report {RR-13-11}, {DCS}, 2013.

\bibitem{rice_algorithm_1976}
John~R. Rice.
\newblock The algorithm selection problem.
\newblock {\em Advances in Computers}, 15:65--118, 1976.

\bibitem{schmee_simple_1979}
Josef Schmee and Gerald~J. Hahn.
\newblock A simple method for regression analysis with censored data.
\newblock {\em Technometrics}, 21(4):417--432, 1979.

\bibitem{witten_data_2011}
Ian~H. Witten, Eibe Frank, and Mark~A. Hall.
\newblock {\em Data Mining: Practical Machine Learning Tools and Techniques}.
\newblock Morgan Kaufmann, 2011.

\bibitem{wolpert_stacked_1992}
David~H. Wolpert.
\newblock Stacked generalization.
\newblock {\em Neural Networks}, 5:241--259, 1992.

\bibitem{wolpert_no_1997}
David~H. Wolpert and William~G. Macready.
\newblock No free lunch theorems for optimization.
\newblock {\em {IEEE} Transactions on Evolutionary Computation}, 1(1):67--82,
  1997.

\bibitem{xu_hierarchical_2007}
Lin Xu, Holger~H. Hoos, and Kevin Leyton-Brown.
\newblock Hierarchical hardness models for {SAT}.
\newblock In {\em {CP}}, pages 696--711, 2007.

\bibitem{xu_hydra_2010}
Lin Xu, Holger~H. Hoos, and Kevin Leyton-Brown.
\newblock Hydra: Automatically configuring algorithms for portfolio-based
  selection.
\newblock In {\em Twenty-Fourth Conference of the Association for the
  Advancement of Artificial Intelligence}, pages 210--–216, 2010.

\bibitem{xu_satzilla_2008}
Lin Xu, Frank Hutter, Holger~H. Hoos, and Kevin Leyton-Brown.
\newblock {SATzilla:} portfolio-based algorithm selection for {SAT}.
\newblock {\em J. Artif. Intell. Res.}, 32:565--606, 2008.

\bibitem{xu_hydra-mip_2011}
Lin Xu, Frank Hutter, Holger~H. Hoos, and Kevin Leyton-Brown.
\newblock Hydra-{MIP:} automated algorithm configuration and selection for
  mixed integer programming.
\newblock In {\em {RCRA} Workshop on Experimental Evaluation of Algorithms for
  Solving Problems with Combinatorial Explosion}, pages 16--30, 2011.

\end{thebibliography}
\bibliographystyle{plain}

\end{document}